\documentclass[a4paper]{llncs}
\usepackage{graphicx}
\usepackage{cite}
\usepackage{subfigure}
\usepackage{amsmath}
\usepackage{amssymb}

\spnewtheorem{ex}[theorem]{Example}{\bfseries}{}%\itshape}

\newcommand{\bit}{BLS}
\newcommand{\Bit}{BLS}
\newcommand{\block}{BTS}
\newcommand{\Block}{BTS}

\begin{document}

\title{Genetic Representations for Evolutionary Minimization of Network Coding Resources}

\author{
Minkyu Kim\inst{1}
\and Varun Aggarwal\inst{2}
\and Una-May O'Reilly\inst{2}
\and \\Muriel M\'{e}dard\inst{1}
\and Wonsik Kim\inst{1}
}

\institute{
Laboratory for Information and Decision Systems
\and
Computer Science and Artificial Intelligence Laboratory\\
Massachusetts Institute of Technology, Cambridge, MA 02139, USA\\
\email{\{minkyu@, varun\_ag@, unamay@csail., medard@, wskim14@\}mit.edu}
}

\maketitle

\begin{abstract}
We demonstrate how a genetic algorithm solves the  problem of minimizing the resources used for network coding, subject to a throughput constraint, in a multicast scenario. A genetic algorithm avoids the computational complexity that makes the problem NP-hard and, for our experiments, greatly improves on sub-optimal solutions of established methods. We compare two different genotype encodings, which tradeoff search space size with fitness landscape, as well as the associated genetic operators. Our finding favors a smaller encoding despite its fewer intermediate solutions and demonstrates the impact of the modularity enforced by genetic operators on the performance of the algorithm.
\end{abstract}

\section{Introduction}
\label{sec:Intro}

\begin{figure}[h]
\vspace{-0.2in}
\centerline{
\subfigure[Network $B$]{\includegraphics[height=1.2in]{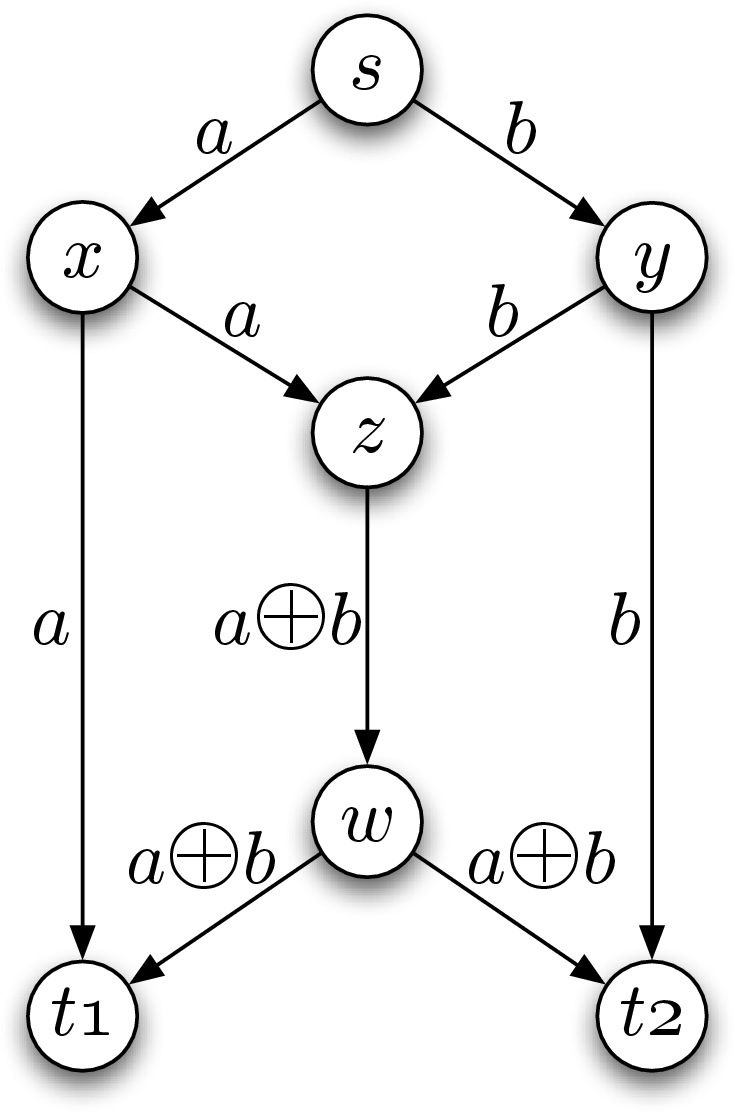}
\label{fig:bf}}
\hfil
\subfigure[Network $B'$]{\includegraphics[height=1.2in]{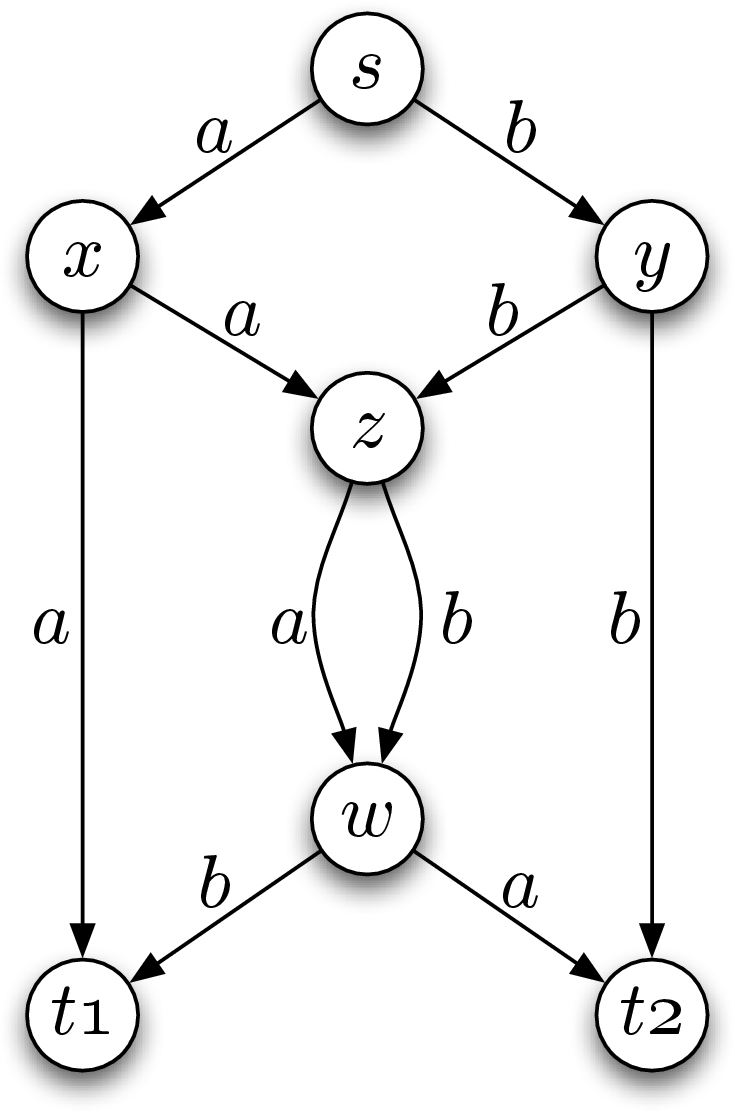}
\label{fig:bf2}}
\hfil
\subfigure[Network $C$]{\includegraphics[height=1.2in]{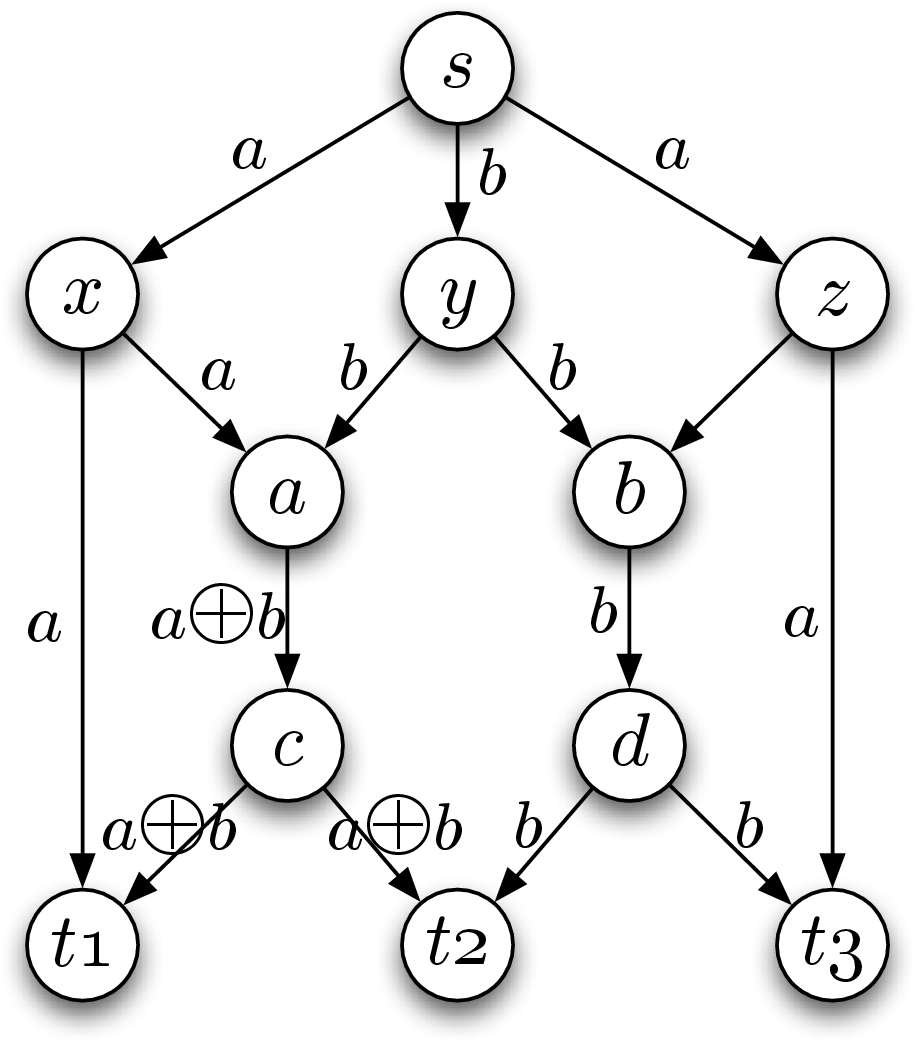}
\label{fig:corr}}
}
\caption{Sample networks for Example \ref{ex:intro}}
\label{fig:sample}
\vspace{-0.2in}
\end{figure}

Network coding is a novel technique that generalizes routing. In traditional routing, each interior network node, which is not a source or sink node, simply forwards the received data or sends out multiple copies of it. In contrast, network coding allows interior network nodes to perform arbitrary mathematical operations, e.g., summation or subtraction, to combine the data received from different links. It is well known that network throughput can be significantly increased by network coding \cite{ACLY00, LYC03, FLW06}. While network coding is assumed to be done at all possible nodes in most of the network coding literature, it is often the case that network coding is required only at a subset of nodes to achieve the desired throughput. Consider Example 1:

\begin{ex}
In the canonical example of network $B$ (Fig. \ref{fig:bf}) \cite{ACLY00}, where each link has unit capacity, source $s$ can send 2 units of data simultaneously to the sinks $t_1$ and $t_2$, which is not possible with routing alone. But only node $z$ needs to combine its two inputs while all other nodes perform routing only. If we suppose that link $(z,w)$ in network $B$ has capacity 2, which we represent by two parallel unit-capacity links in network $B'$ (Fig. \ref{fig:bf2}), a multicast of rate 2 is possible without network coding. In network $C$ (Fig. \ref{fig:corr}), where node $s$ is to transmit data at rate 2 to the 3 leaf nodes, network coding is required either at node $a$ or at node $b$, but not at both.
\hfill $\square$
\label{ex:intro}
\end{ex}

Example~1 leads us to the following question: To achieve the desired throughput, at which nodes does network coding need to occur? This question's answer is valuable because eliminating unnecessary coding nodes will save computation at the application layer if that is where network coding is handled. Alternatively, if network coding is integrated in the buffer management of routers, it will reduce the number of routers that need to perform coding operations without compromising communication capacity.  For a GA, the problem can be posed as the minimization of coding cost (in links or nodes) subject to the constraint of feasibility (achieving the desired throughput).

The problem of determining a minimal set of nodes where coding is required is NP-hard; its decision problem, which decides whether the given multicast rate is achievable without coding, reduces to a multiple Steiner subgraph problem, which is NP-hard~\cite{RP86}. It is shown that even approximating the minimum number of coding nodes within any multiplicative factor or within an additive factor of $|V|^{1-\epsilon}$ is NP-hard \cite{LSB05}. Note, however, that once the set of coding nodes is identified, a network code achieving the desired throughput can be efficiently constructed for the multicast scenario, either in a deterministic \cite{JSC05} or randomized fashion \cite{HKM03}.

In the network research community, \cite{KAME06} and \cite{KMAO07} have documented results that demonstrate the benefit of the GA over other existing approaches in terms of reducing the number of coding links or nodes and its applicability to a variety of generalized scenarios. These contributions emphasized the computational ``how-to'' aspects of feasibility checking and the transformations of the network graph into secondary graphs that express possible coding situations uniformly since these are key to evaluating the fitness function of the GA.

In the course of investigating the feasibility checks, graph transformations, and value of using a GA, we experimented with two different genotype encodings\footnote{To minimize confusion, throughout the paper, the term ``encoding" refers to ``genotype encoding" only, while the term ``coding" means ``network coding."} and associated operators. For both encodings, we use a genotype composed of a number of blocks, each of which consists of a set of variables indicating the link states. For a block of length $k$, using an alphabet of cardinality 2, the \emph{Binary Link State} (\bit{}) encoding represents all possible $2^k$ states of $k$ links. On the other hand, for the \emph{Block Transmission State} (\block{}) encoding, we group those link states into $(k+2)$ transmission states. Despite the smaller search space size of \block{} encoding, it is not clear that it should be superior to \bit{} encoding because in grouping many link states into one, less information that would relate the fitness of solutions {\it intermediate} to the best solution is available in contrast to \bit{} encoding which provides more information through its intermediate solutions.

In this paper we focus on the two different encodings with associated genetic operators and conduct a more comprehensive comparison between them. Specifically relevant to the GA community, we consider into the GA encoding tradeoff issues related to search space size and fitness landscape. The rest of the paper is organized as follows. Section~\ref{sec:problem} presents the problem formulation, and Section~\ref{sec:NCGA} describes the network coding GA (NCGA) with the two different encodings and associated  operators. Section~\ref{sec:experiments} sets up a set of experiments into relative values of the encodings and discusses the results. Section \ref{sec:con} presents a summary of the results and our conclusions.

\section{Problem Formulation}
\label{sec:problem}

We assume that a network is given by a directed multigraph $G=(V,E)$ as in \cite{KM03} where each link has a unit capacity whose unit can be arbitrarily chosen, e.g., $k$ bits per second for a constant $k$, or a fixed size packet per unit time, etc. Links with larger capacities are represented by multiple links. Only integer flows are allowed, hence there is either no flow or a unit rate of flow on each link. We consider the single multicast scenario in which a single source $s \in V$ wishes to transmit data at rate $R$ to a set $T \subset V$ of sink nodes. Rate $R$ is said to be achievable if there exists a transmission scheme that enables all $|T|$ sinks to receive all of the information sent. We only consider linear coding, where a node's output on an outgoing link is a linear combination of the inputs from its incoming links. Linear coding is sufficient for multicast~\cite{LYC03}.

Given an achievable rate $R$, we wish to determine a minimal set of nodes where coding is required in order to achieve this rate. However, whether coding is necessary at a node is determined by whether coding is necessary at at least one of the node's outgoing links and thus, as pointed out also in \cite{LSB05}, the number of coding links is in fact a more accurate estimator of the amount of computation incurred by coding. We assume hereafter that our objective is to minimize the number of coding \emph{links} rather than \emph{nodes}. Note, however, that as demonstrated in \cite{KAME06}, it is straightforward to generalize the proposed algorithm to the case of minimizing the number of coding nodes. Furthermore, \cite{KAME06} shows that, with appropriate changes, the algorithm can be readily applied to more generalized optimization scenarios, e.g., when links and nodes have different coding costs.

It is clear that no coding is required at a node with only a single input since these nodes have nothing to combine with \cite{KAME06}. For a node with multiple incoming links, which we refer to as a \emph{merging node}, if the linearly coded output to a particular outgoing link weights all but one incoming message by zero, effectively no coding occurs on that link; even if the only nonzero coefficient is not identity, there is another coding scheme that replaces the coefficient by identity~\cite{LSB05}. Thus, to determine whether coding is necessary at an outgoing link of a merging node, we need to verify whether we can constrain the output of the link to depend on a single input without destroying the achievability of the given rate. As in network $C$ of Example \ref{ex:intro}, the necessity of coding at a link depends on which other links code and thus the problem of deciding where to perform network coding in general involves a selection out of exponentially many possible choices. We employ a GA-based search method to efficiently address the large and exponentially scaling size of the space.

\section{Network Coding GA (NCGA)}\label{sec:NCGA}

Prior to using the NCGA, the given network graph $G$ is transformed into a secondary graph by either of the two methods presented in \cite{KAME06, KMAO07}\footnote{An interested reader is referred to \cite{KAME06, KMAO07} for the details of the two methods for graph transformation and feasibility testing.}. Regardless which method is used, mapping the network coding problem to a GA framework is done as follows.

\begin{figure}[h]
\vspace{-0.2in}
\centerline{
\subfigure[Merging node $v$]{\includegraphics[height=1.4in]{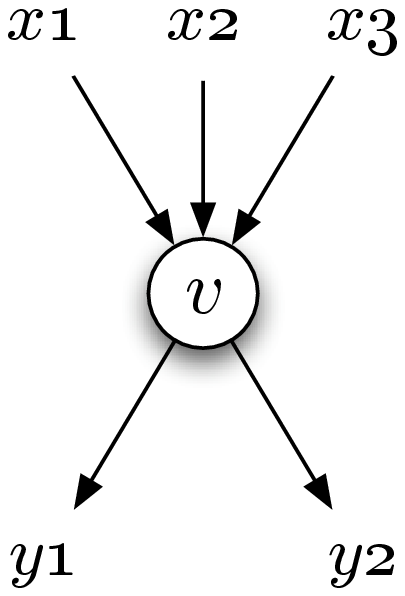}
\label{fig:block1}}
\hfil
\subfigure[Two blocks for outgoing links of $v$]{\includegraphics[height=1.4in]{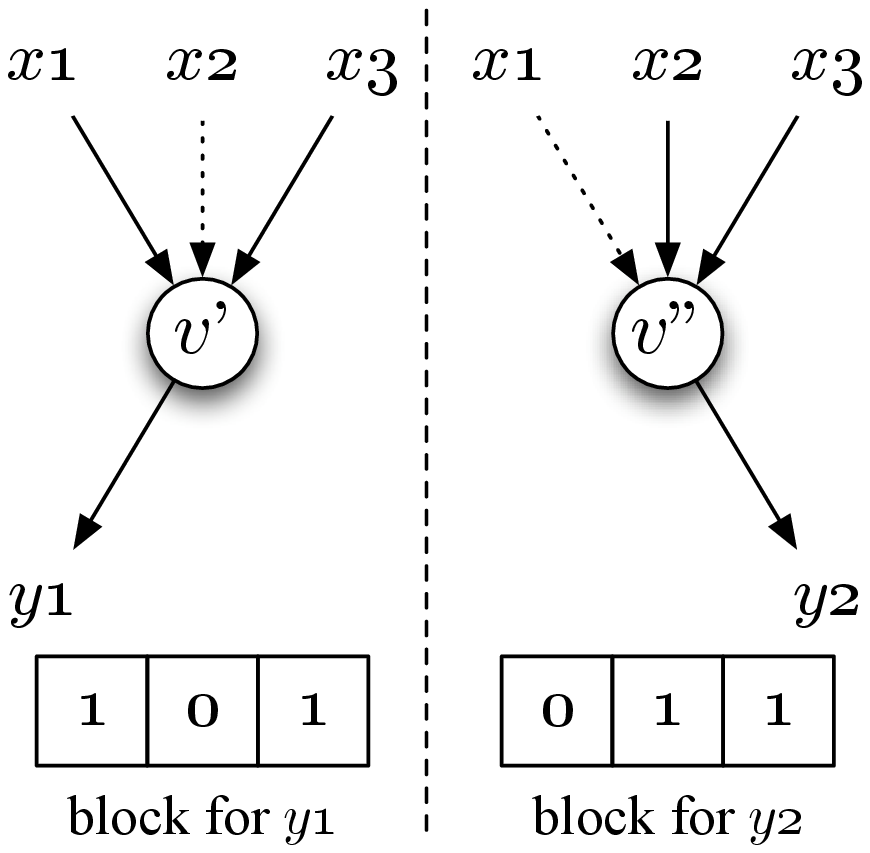}
\label{fig:block2}}
}
\caption{Node $v$ with 3 incoming and 2 outgoing links results in 2 blocks, each with 3 variables indicating the states of incoming links $(x_1, x_2, x_3)$ onto the associated outgoing link.}
\label{fig:block}
\vspace{-0.2in}
\end{figure}

Suppose a merging node with $k (\geq 2)$ incoming links. To consider the transmission to \emph{each} of its outgoing links, we assign a binary variable to each of its $k$ incoming links, which being 1 indicates that the link state is \emph{active} (the input from the associated incoming link is transmitted to the outgoing link) and 0 indicates it is \emph{inactive}. Given that network coding is required for the transmission only if two or more link states are active, we may need to consider those $k$ variables together. We refer to the set of the $k$ variables as a \emph{block} of length $k$ (see Fig. \ref{fig:block} for an example). The way how those binary variables are actually encoded as a genotype will be described later in this section.

\noindent {\bf Constraint and Fitness Function} A genotype is called \emph{feasible} if there exists a network coding scheme that achieves the given rate $R$ with the link states determined by the genotype. To calculate the fitness of genotype $\underline{y}$, its feasibility must be checked by either of the two methods in \cite{KAME06}, \cite{KMAO07} depending on the secondary graph chosen earlier, and the fitness value $F$ is assigned as
\begin{equation*}
F(\underline{y})=
    \begin{cases}
        \text{number of blocks with two or more active links}, & \text{if } \underline{y} \text{ is feasible,} \\
        \infty, & \text{if } \underline{y} \text{ is infeasible}.
    \end{cases}
\end{equation*}

The NCGA uses a standard generation-based GA control loop with tournament selection. It terminates at some maximum number of generations. Afterward, the best solution of the run is optimized with \emph{greedy sweep}: we switch each of the remaining 1's to 0 if it can be done without violating feasibility. This procedure may only improve the solution, and sometimes the improvement can be substantial. Reference \cite{KMAO07} proves that the NCGA with greedy sweep is guaranteed to perform no worse than the existing algorithm in \cite{LSB05}.

%\subsection{Genotype Encodings}\label{sec:encodings}

\subsubsection{Binary Link State (\Bit{}) Encoding and Operators}\label{sec:BLS}

This encoding allows a block of length $k$ to take any of $2^k$ possible binary strings of length $k$. If we denote by $d^v_{in}$ and $d^v_{out}$ the in-degree and the out-degree of node $v$, node $v$ has $d^v_{out}$ blocks of length $d^v_{in}$, and we have a total of $m=\sum_{v \in \mathcal{V}} d^v_{in} d^v_{out}$ binary variables, where $\mathcal{V}$ is the set of all merging nodes. We must explore the $m$-dimensional binary space of $2^{m}$ candidates to find the desired minimal set of coding links.

For \bit{} encoding we use uniform crossover, where each pair of genotypes is selected for crossover with a given probability (mixing ratio) and the two genotypes in a selected pair exchange each bit independently with another given probability (crossover probability). For mutation, we use simple binary mutation, where each bit in each genotype is flipped independently with a given probability (mutation rate). Since these operators deal with each bit separately, we refer to the operators used for \bit{} encoding as \emph{bit-wise} genetic operators.

\subsubsection{Block Transmission State (\Block{}) Encoding and Operators}\label{sec:BTS}

As mentioned above, once a block has at least two 1's, replacing all the remaining 0's with 1's has no effect on whether coding is done. Moreover, it can be shown that substituting 0 with 1, as opposed to substituting 1 with 0, does not hurt the feasibility. Therefore, for a feasible genotype, any block with two or more 1's can be treated the same as the block with all 1's. Thus we could group all the states with two or more active links into a single state, {\it coded} transmission. This state is rounded out by $k$ states for the {\it uncoded} transmissions of the input received from one of the $k$ single incoming links and one state indicating {\it no} transmission. Thus \block{} encoding emerges where each block of length $k$ can only take one of the following $(k+2)$ strings: $``111...1"$, $``100...0"$, $``010...0"$, $``001...0"$, $...$, $``000...1"$, $``000...0"$. The net effect is a reduction in the number of possible states for a block to $(k+2)$ rather than $2^{k}$. If we let $w$ be the total number of blocks (i.e., $w=\sum_{v \in \mathcal{V}} d^v_{out}$) and $k_i$ denote the length of the $i$-th block $(i=1, ..., w)$, the search space size is $\prod_{i=1}^{w} (k_i+2)$. However, the benefit of the smaller space size in fact comes at the price of losing the information on the partially active link states that may serve as intermediate steps toward an uncoded transmission state. This tradeoff will be discussed more in depth in Section \ref{sec:experiments}.

To preserve the \Block{} encoding structure throughout genetic operations, we need to define a new set of genetic operators, which we refer to as \emph{block-wise} genetic operators. For block-wise uniform crossover, we let two genotypes subject to crossover exchange each block, rather than bit, independently with the given crossover probability. For block-wise mutation, we let each block under mutation take another string chosen uniformly at random out of $(k+1)$ other strings for a length-$k$ block. If mutation rate is $\alpha$, the average number of changed bits in a length-$k$ block is now calculated as $\frac{4k^2}{(k+1)(k+2)} \alpha$, whereas it is $k \alpha$ for bit-wise mutation. While the difference becomes more apparent when $k$ is large since the latter is upper bounded by $4\alpha$, those values are still different for $k=2$, where the block structure makes no difference in the space size. Though block-wise mutation may lead to much smaller number of flipped bits, it is more likely to cause a sudden change in a genotype.

\section{Experiments}\label{sec:experiments}

\subsection{Experiment Setup}

The two encodings not only differ in the size of search space, but in the way the genetic operators are applied. In \block{} encoding with the block-wise operators, crossover is applied at the block boundaries and mutation is performed intra-block. However, in \bit{} encoding with the bit-wise operators, crossover and mutation are randomly applied without respecting any block boundaries. While evaluating the effect of the search space size reduction, we also want to investigate whether the exploitation of block level modularity by the block-wise operators gives any significant improvement in the algorithm's performance. We thus set up two experiments: Experiment I compares the effect of the two encodings combined with associated operators on the performance of the NCGA, while Experiment II tests the effect of the operators alone by isolating the effect of the encodings that lead to different space sizes.

\textbf{Experiment I}: We use two acyclic networks, I-50 and I-75, generated by the algorithm in~\cite{MP04}, whose details are given in Table %~\ref{networkdetails}
1. Note that \block{} encoding reduces the size of the search space by 30.3 and 115 orders of magnitude for networks I-50 and I-75, respectively, compared with that in the case of \bit{} encoding. This experiment tests which encoding is better given the tradeoff in the search space size and ease of traversing the fitness landscape.

\textbf{Experiment II}: We construct a set of synthetic networks with only blocks of length 2. Note that for a block of length 2, the two encodings have the same search space size ($2^k = k+2$ when $k=2$), but the block-wise operators retain their modularity. These networks are constructed by cascading a number of copies of network $B'$ in Fig. \ref{fig:bf2} such that the source of each subsequent copy of $B'$ is replaced by an earlier copy's sink. We use fixed-depth binary trees containing 3, 7, 15, and 31 copies of $B'$ (henceforth called II-3, II-7, II-15, and II-31, respectively). Parameters of these networks are given in Table %~\ref{tab:networkdetails}.
1. All these network have 0 as the minimum number of coding links, i.e., multicast rate 2 is achievable without coding.  We scale up the network size to investigate the payoff one gets with modular operators as the search space size increases.

We use the NCGA with the decomposition-based graph transformation and the max-flow feasibility testing described in \cite{KMAO07}. For comparison, we also perform experiments using the two existing approaches by Fragouli et al.~\cite{FS06} (``Minimal 1")\footnote{Though minimizing network coding resources is not its main concern, \cite{FS06} presents an algorithm to obtain a subgraph with a minimal number of coding links.}, and Langberg et al.~\cite{LSB05} (``Minimal 2"), in both of which link removal is done in a random order. For Minimal 1, the subgraph is selected also by a minimal approach, which starting from the original graph sequentially removes the links whose removal does not destroy the achievability.
\begin{table}[h]
\vspace{-0.15in}
\centering{
\begin{tabular} {|c|c|c|c|c|}
\hline
Network & Genotype & Number of & Avg. length & Search space size ($\log_{10}$)\\
          &  length     & blocks         &   of blocks &  (BLS/BTS)\\
\hline
\hline
I-50 & 280 & 71 & 3.94 & 84.29/53.93\\
\hline
I-75 & 761 & 130 & 5.85 & 229.08/113.47\\
\hline
\hline
II-3\,\,\, & 32 & 16 & 2 & 9.63/9.63\\
\hline
II-7\,\,\, & 80 & 40 & 2 & 24.08/24.08\\
\hline
II-15 & 176 & 88 & 2 & 52.98/52.98\\
\hline
II-31 & 368 & 184 & 2 & 110.78/110.78\\
\hline
\end{tabular}
\caption{Details of the networks used in the experiments.}}
\vspace{-0.5in}
\label{tab:networkdetails}
\end{table}

\subsection{Algorithm Parameters}

We set the total budget of fitness evaluations to 150,000 (a very small fraction of the search space size of the networks considered. Preliminary experiments suggested different tournament sizes and mutation rates for the two encodings: 10 and 0.006 for \bit{} encoding, and 100 and 0.012 for \block{} encoding, respectively. All other parameters are matched for the two encodings. We perform 30 runs for each network with both encodings. Table %\ref{tab:gaparams}
2 summarizes the parameters used.
\begin{table}[h]
\vspace{-0.15in}
\centering{
\begin{tabular}{|l|c|}\hline
Population size & 150  \\ \hline
Tournament size & 10(BLS)/100(BTS) \\ \hline
Maximum generations & 1000 \\ \hline
Mixing ratio/Crossover probability &  0.8/0.8\\ \hline
Mutation rate &  0.006(BLS)/0.012(BTS) \\ \hline
%Number of runs & 30 \\ \hline
\end{tabular}
\caption{GA parameters used in the experiments.}}
\vspace{-0.3in}
\label{tab:gaparams}
\end{table}

When randomly initializing the population, we insert an all-one vector, which represents the solution where coding is done everywhere and thus is feasible by the assumption that the given rate $R$ is achievable. The role of the all-one vector as a feasible starting point is crucial to the performance of the algorithm as discussed in \cite{KAME06}.

\subsection{Experimental Results}
Results for the both experiments are summarized in Table %~\ref{resultsummary}
3. The table shows the optimal fitness achieved by each algorithm averaged over 30 runs. The statistical significance of the difference between \block{} and \bit{} encodings is measured by conducting paired $t$-tests and the $p$-values are reported in the last row of the table.

\begin{table}[h]
\vspace{-0.15in}
\centering{
\begin{tabular} {|c|r|r|r|r|r|r|r|}\hline
& \multicolumn{1}{|c|}{I-50} & \multicolumn{1}{|c|}{I-75} && \multicolumn{1}{|c|}{II-3} & \multicolumn{1}{|c|}{II-7} & \multicolumn{1}{|c|}{II-15} & \multicolumn{1}{|c|}{II-31} \\ \hline
NCGA/\Bit{} & 3.33(1.03) & 6.43(1.30) && 0.93(0.69) & 2.20(1.27) & 5.57(1.55) & 12.43(2.37)\\ \cline{2-8}
{\footnotesize(w/o greedy sweep)} & 3.33(1.03) & 39.93(2.74) && 0.93(0.69) & 2.20(1.27) & 5.57(1.55) & 12.43(2.37)\\ \hline
NCGA/\Block{} & 2.40(0.62) & 3.63(0.61) && 0.00(0.00) & 0.00(0.00) & 0.17(0.38) & 1.03(0.81)\\  \cline{2-8}
{\small(w/o greedy sweep)} & 2.40(0.62) & 3.63(0.61) && 0.00(0.00) & 0.00(0.00) & 0.17(0.38) & 1.07(0.83)\\  \hline
Minimal 1  & 4.90(1.37) & 9.50(2.16) && 3.00(0.00) & 7.00(0.00)& 15.50(0.00) & 31.00(0.00)\\ \hline
Minimal 2  & 4.33(1.37) & 7.90(1.71) && 2.13(0.86)&  4.37(1.25) & 9.90(1.65) & 19.97(2.66) \\ \hline
$p$-value &  \multicolumn{1}{|c|}{$8.21e^{-5}$} & \multicolumn{1}{|c|}{$2.65e^{-15}$} && \multicolumn{1}{|c|}{$6.7e^{-10}$} & \multicolumn{1}{|c|}{$2.2e^{-13}$} & \multicolumn{1}{|c|}{$2.9e^{-26}$} & \multicolumn{1}{|c|}{$1.55e^{-32}$} \\ \hline
\end{tabular}
\caption{Performance of the algorithms for each network. Each value in brackets is standard deviation.}}
\label{resultsummary}
\vspace{-0.3in}
\end{table}

\textbf{Experiment I}: For both networks I-50 and I-75, the NCGA with greedy sweep, with either of the two encodings, outperforms the two existing minimal approaches. Between the two encodings, \block{} encoding gives rise to a substantial performance gain over \bit{} encoding with the statistical significance confirmed by the tabulated p-values.
%\bit{} NCGA performance detoriates significantly for I-75 without the greedy sweep (average optimal solution: 39.93 coding links) in comparison to \block{} (average optimal solution: 5.67). This result is important because greedy sweep is not possible in the distributed version of the genetic algorithm \cite{KMAO07}.

\textbf{Experiment II}: Again the NCGA with \block{} encoding outperforms that with \bit{} encoding on average for all networks, while either of the two performs significantly better than the minimal algorithms. For networks II-3 and II-7, the NCGA with \block{} encoding finds the optimal (0 coding links) in all of the 30 runs. For networks II-15 and II-31, it succeeds to find the optimal solution 25 and 8 times, respectively. On the other hand, \bit{} encoding does not find the optimal number of coding links in any of the 30 runs for networks II-7, II-15, and II-31.

\begin{figure}[h]
\vspace{-0.2in}
    \centering
    \includegraphics[height=1.8in]{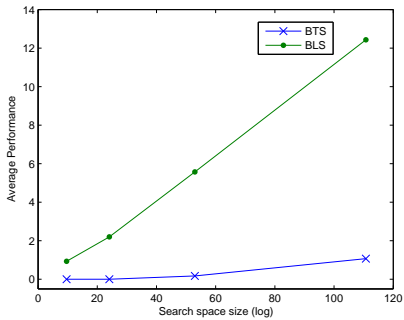}
    \caption{Average performance of NCGAs with $\log$ of search space size.}
\label{graph}
\vspace{-0.2in}
\end{figure}

The average performance of the NCGA with both encodings is plotted against the logarithm of the search space size in Fig.~\ref{graph}. The plot suggests a linear scaling of algorithms as the search space size grows exponentially. More data points would lend more confidence to this hypothesis. The curve for \block{} encoding has a much smaller intercept and slope than \bit{} encoding, implying that the payoff of the block-wise operators increases as the search space size increases.

%\vspace{-0.3in}
\subsection{Discussion of Results}

Experiment I clearly indicates that \block{} encoding is better than \bit{} encoding for the networks considered. We can thus conclude that the benefits of the smaller search space trump the challenge of the more difficult fitness landscape. For network I-50,  \block{} encoding improves over \bit{} encoding on average by a single coding link. Though small, this difference is statistically significant. For network I-75, without greedy sweep, the average difference in performance between the two algorithms is much higher, i.e., 34 coding links.  This large difference in performance can be attributed to two specific factors: the much larger search space size (see Table %~\ref{tab:networkdetails}
1) and larger average block size. The difference also indicates that the information on the intermediate solutions that \bit{} encoding provides may not be particularly useful without guaranteeing that those intermediate steps ultimately lead to an uncoded transmission state. 

Experiment II demonstrates the superiority, by a remarkably large margin, of the block-wise operators over the bit-wise operators. It also indicates that both NCGAs scale linearly with an exponentially growing search space size (see Fig.~\ref{graph}), which is remarkable. This prompts due analysis of the difference between the two operators. When applied to the pair of blocks $``00"$ and $``11"$, the block-wise crossover cannot result in either block $``01"$ or $``10"$. However, for the bit-wise crossover, the pair of blocks $``00"$ and $``11"$ may result in $``00"$, $``01"$, $``10"$, or $``11"$. It can be shown that with probability $\frac{1}{4}$ the two crossovers behave differently, if the population has equal frequency of all block types. Let us recall that the block-wise mutation leads to a smaller number of changed bits on average than the bit-wise mutation (Section \ref{sec:NCGA}). Nevertheless, the block-wise mutation exhibits higher ``exploratory power" than the bit-wise mutation in the sense that it is more likely to lead to changes in multiple bits. For the block-wise mutation, given any block, the remaining three blocks are equally likely to occur on mutation. Thus, if mutation rate is $\alpha$, the probabilities of 0, 1, 2-bit change are $1-\alpha$, $\frac{2}{3}\alpha$, $\frac{1}{3}\alpha$, respectively, whereas those probabilities in the bit-wise case are $(1-\alpha)^2$, $2\alpha(1-\alpha)$, $\alpha^2$, respectively. Provided that $\alpha < \frac{1}{3}$, the probability of 2-bit change is larger for the block-wise mutation. A similar analysis can be done for the whole genotype as well.

One may speculate that the better performance of the block-wise operators is due to the higher exploratory power of the block-wise mutation rather than the modularity of the operators. To confirm the contrary, we consider a new set of operators, called the \emph{ Matched Hamming Distance} (MHD) operators, where the MHD mutation leads to the statistically same Hamming distance changes as the block-wise mutation, but exhibits no positional bias as to where the mutation is applied, and the MHD crossover is the same as the bit-wise crossover which neither imposes modularity. From Table %~\ref{resultstest}
4 compared with Table 3, we observe that the MHD operators perform similarly as the bit-wise operators, but far worse than the block-wise operators. We can thus confidently claim that the respect for modularity enforced by the block-wise operators is the main cause of the superior performance of the block-wise operators.

\begin{table}[h]
\vspace{-0.15in}
\centering{
\begin{tabular} {|c|c|c|c|c|}\hline
& II-3 & II-7 & II-15 & II-31 \\ \hline
NCGA/MHD & 0.77(0.68) & 2.47(1.33) & 5.83(1.68) & 12.63(3.23)\\ \hline
%Bit-wise & 2.20(1.27) & 5.57(1.55) \\ \hline
%Block-wise & 0.00(0.00) & 0.17(0.38) \\ \hline
\end{tabular}
\caption{Performance of NCGA with MHD operators. Refer to Table 3 for comparison with NCGAs with bit-wise or block-wise operators.}}
\vspace{-0.3in}
\label{resultstest}
\end{table}

\section{Conclusions and Future Work} \label{sec:con}
For our suite of network coding problems, we have found that the benefits of the smaller search space and modular operators trump the challenge of the more difficult fitness landscape. In the future, we will study the effect of exploiting further modularity with BTS operators that cross over at merging node boundaries and perform intra-block mutations. We could incorporate hierarchical modularity using domain knowledge of sparsely connected regions, regions with similar structure or simply neighboring nodes. A hierarchical structure of crossover boundaries could be formed and applied with different probabilities. These results will inform the GA community and help push the state-of-art in algorithms for minimizing network coding resources.

\bibliographystyle{splncs}
\bibliography{EvoComnet07}

\begin{thebibliography}{10}

\bibitem{ACLY00}
Ahlswede, R., Cai, N., Li, S.Y.R., Yeung, R.W.:
\newblock Network information flow.
\newblock {IEEE} Trans. Inform. Theory \textbf{46}(4) (2000)  1204--1216

\bibitem{LYC03}
Li, S.Y.R., Yeung, R.W., Cai, N.:
\newblock Linear network coding.
\newblock {IEEE} Trans. Inform. Theory \textbf{49}(2) (2003)  371--381

\bibitem{FLW06}
Fragouli, C., {Le Boudec}, J.Y., Widmer, J.:
\newblock Network coding: An instant primer.
\newblock SIGCOMM Comput. Commun. Rev. \textbf{36}(1) (2006)  63--68

\bibitem{RP86}
Richey, M.B., Parker, R.G.:
\newblock On multiple {S}teiner subgraph problems.
\newblock Networks \textbf{16}(4) (1986)  423--438

\bibitem{LSB05}
Langberg, M., Sprintson, A., Bruck, J.:
\newblock The encoding complexity of network coding.
\newblock In: Proc. {IEEE} ISIT. (2005)

\bibitem{JSC05}
Jaggi, S., Sanders, P., Chou, P.A., Effros, M., Egner, S., Jain, K., Tolhuizen,
  L.:
\newblock Polynomial time algorithms for multicast network code construction.
\newblock {IEEE} Trans. Inform. Theory \textbf{51}(6) (2005)  1973--1982

\bibitem{HKM03}
Ho, T., Koetter, R., M\'{e}dard, M., Karger, D.R., Effros, M.:
\newblock The benefits of coding over routing in a randomized setting.
\newblock In: Proc. {IEEE} ISIT. (2003)

\bibitem{KAME06}
Kim, M., Ahn, C.W., M\'{e}dard, M., Effros, M.:
\newblock On minimizing network coding resources: An evolutionary approach.
\newblock In: Proc. NetCod. (2006)

\bibitem{KMAO07}
Kim, M., M\'{e}dard, M., Aggarwal, V., O'Reilly, U.M., Kim, W., Ahn, C.W.,
  Effros, M.:
\newblock Evolutionary approaches to minimizing network coding resources.
\newblock In: Proc. IEEE Infocom (to appear). (2007)

\bibitem{KM03}
Koetter, R., M\'{e}dard, M.:
\newblock An algebraic approach to network coding.
\newblock {IEEE/ACM} Trans. Networking \textbf{11}(5) (2003)  782--795

\bibitem{MP04}
Melan\c{c}on, G., Philippe, F.:
\newblock Generating connected acyclic digraphs uniformly at random.
\newblock Inf. Process. Lett. \textbf{90}(4) (2004)  209--213

\bibitem{FS06}
Fragouli, C., Soljanin, E.:
\newblock Information flow decomposition for network coding.
\newblock {IEEE} Trans. Inform. Theory \textbf{52}(3) (2006)  829--848

\end{thebibliography}

\end{document}